\documentclass{scrartcl} 
\usepackage{authblk}

\usepackage{url}
\usepackage{float}
\floatstyle{plaintop}
\restylefloat{table}
\usepackage{graphicx}
\usepackage[labelfont=bf, ]{caption}

\begin{document}
\title{How Many Annotators Do We Need?}
\subtitle{A Study on the Influence of Inter-Observer Variability on the Reliability of Automatic Mitotic Figure Assessment}

\author[1\thanks{equal contribution}]{Frauke~Wilm} 
\author[2$^\ast$]{Christof~A.~Bertram} 
\author[1]{Christian~Marzahl} 
\author[3]{Alexander~Bartel} 
\author[4]{Taryn~A.~Donovan} 
\author[5]{Charles-Antoine~Assenmacher} 
\author[6]{Kathrin~Becker} 
\author[7]{Mark~Bennett} 
\author[8]{Sarah~Corner} 
\author[9]{Brieuc~Cossic} 
\author[10]{Daniela~Denk} 
\author[11]{Martina~Dettwiler} 
\author[7]{Beatriz~Garcia~Gonzalez} 
\author[11]{Corinne~Gurtner} 
\author[12]{Annabelle~Heier} 
\author[12]{Annika~Lehmbecker} 
\author[2,12]{Sophie~Merz} 
\author[7]{Stephanie~Plog} 
\author[12]{Anja~Schmidt} 
\author[12]{Franziska~Sebastian} 
\author[8]{Rebecca~C.~Smedley} 
\author[13]{Marco~Tecilla} 
\author[8]{Tuddow~Thaiwong} 
\author[1]{Katharina~Breininger} 
\author[8]{Matti~Kiupel} 
\author[1]{Andreas~Maier} 
\author[2]{Robert~Klopfleisch} 
\author[1,14]{Marc~Aubreville} 

\affil[1]{Pattern Recognition Lab, Computer Sciences, Friedrich-Alexander-Universit\"at Erlangen-N\"urnberg, Erlangen, Germany}
\affil[2]{Institute of Veterinary Pathology, Freie Universit\"at Berlin, Germany}
\affil[3]{Institute for Veterinary Epidemiology and Biostatistics, Freie Universit\"at Berlin}
\affil[4]{Department of Anatomic Pathology, Animal Medical Center, New York, USA}
\affil[5]{Department of Pathobiology, University of Pennsylvania, Philadelphia, USA}
\affil[6]{Department of Pathology, University of Veterinary Medicine Hannover, Germany}
\affil[7]{Synlab’s VPG Histology, Bristol, UK}
\affil[8]{Veterinary Diagnostic Laboratory, Michigan State University, Lansing, USA}
\affil[9]{Pharmacology \& Preclinical Development, Idorsia Pharmaceuticals Ltd, Switzerland}
\affil[10]{International Zoo Veterinary Group, Keighley, UK}
\affil[11]{Institute of Animal Pathology, Vetsuisse Faculty, University of Bern, Switzerland}
\affil[12]{IDEXX Vet Med Labor GmbH, Kornwestheim, Germany}
\affil[13]{F. Hoffmann-La Roche Ltd, Basel, Switzerland}
\affil[14]{Technische Hochschule Ingolstadt, Ingoldstadt, Germany}
\date{}

\maketitle
\vspace{-5em}
\begin{center}
\url{frauke.wilm@fau.de}
\end{center}

\begin{abstract}
\noindent \textbf{Abstract.} Density of mitotic figures in histologic sections is a prognostically relevant characteristic for many tumours. Due to high inter-pathologist variability, deep learning-based algorithms are a promising solution to improve tumour prognostication. Pathologists are the gold standard for database development, however, labelling errors may hamper development of accurate algorithms. In the present work we evaluated the benefit of multi-expert consensus (n = 3, 5, 7, 9, 11) on algorithmic performance. While training with individual databases resulted in highly variable F$_1$~scores, performance was notably increased and more consistent when using the consensus of three annotators. Adding more annotators only resulted in minor improvements. We conclude that databases by few pathologists and high label accuracy may be the best compromise between high algorithmic performance and time investment.    
\end{abstract}

\section{Introduction}
Histologic examination of tumour specimens is used to derive important information with regards to patient prognosis and selection of appropriate treatment. For numerous tumour types, including canine mast cell tumours, cellular proliferation is one of the most meaningful prognostic parameters. As part of the recommended grading schemes, cells undergoing division (mitotic figures) must be counted in histologic sections. However, identification of mitotic figures has a high degree of inter-observer variability due to inconsistent classification of mitotic figures (as opposed to mitotic-like impostors) or overlooking/omitting mitotic figure candidates~\cite{3023-01,3023-02}. In order to improve reproducibility and accuracy of enumerating mitotic figures, promising deep learning-based algorithms for the automated analysis of digitised histologic sections have been developed~\cite{3023-01,3023-03,3023-04,3023-05}. However, as pathologists are the gold standard for dataset development, visual and cognitive limitations of human experts may hamper the consistency of datasets and subsequently algorithmic performance~\cite{3023-02}.

All available datasets on mitotic figures from human and canine tumours have used not more than two pathologists as annotators for initial labelling in histologic images ~\cite{3023-02,3023-03,3023-04,3023-05,3023-06}. Disagreement between the labels of these two pathologists was reported in up to 68.2~\%~\cite{3023-03}. Divergent labels between these two pathologists were reviewed for final consensus by one~\cite{3023-05} or two~\cite{3023-03,3023-04} additional pathologists or by reassessment of the same experts~\cite{3023-02,3023-06}. Although consensus by multiple pathologists is expected to counterbalance the high inter-rater variability, influence on algorithmic performance has not been examined to date. 

This study aims to evaluate the ideal number of expert opinions required for the development of deep learning-based mitotic figure detection algorithms with high accuracy. For this purpose, a well established object detection algorithm was trained with labels derived from a consensus of a range of pathologists (n~=~1-11) and evaluated against reference annotations derived from the total consensus of twelve pathologists.

\section{Material and methods}
For this investigation, datasets were created from 50 histologic images of canine mast cell tumours from 50 patients. Use of these samples was approved by the local governmental authorities (State Office of Health and Social Affairs of Berlin, approval ID: StN 011/20). Histologic sections were created with routine methods (haematoxylin and eosin stain) and digitised at a resolution of 0.25~${\mu m}$ per pixel (400~x magnification) using an Aperio ScanScope CS2 (Leica, Germany) scanner. 

\subsection{Creation of databases}
Independent databases were created by twelve veterinary pathologists (each at least three years of experience in histopathology) using the SlideRunner 
annotation software~\cite{3023-09}.
For each of the 50 slides, a field of interest in the tumour area with a standard size of $2.37~{mm}^2$ was selected and annotators marked centroid coordinates of all mitotic figures recognised in these tumour regions. Each pathologist identified between 1,324 and 4,412 mitotic figures (total number of annotations:  32,917). The dataset was split into 35 training, 5 validation and 10 test images with similar variability in mitotic figure density in the selected regions. 

\subsection{Deep learning-based mitotic figure detection}
For mitotic figure detection we customised a publicly available RetinaNet implementation~\cite{3023-01} with a ResNet18 stem. For network training, 2,500 patches (1,024~x~1,024 pixels at highest resolution) each containing at least one mitotic figure were randomly drawn from the 35 training images. The network was trained with a variable number of databases created by the pathologists. The training and validation reference was individually defined as the majority vote of this subgroup. The training was split into two phases. First, only the randomly initialised network heads were trained (batch size: 12) for five epochs using a maximal learning rate of $10^{-3}$. Afterwards, the complete model was trained for an additional ten epochs and a maximal learning rate of $10^{-4}$. During this second phase, 1,500 patches from the validation set were used for model selection.

Four different training set-ups were evaluated: (1) The network was trained with each individual pathologist's database and the model performance was compared to the annotator's performance on the test set. To ensure stability, the F$_1$~score was computed as median of three independent training runs. (2) The network was trained on the consensus of an increasingly larger, randomly chosen subset of pathologists. In order to obtain unambiguous agreement, these increases consisted only of odd increments. For each addition, ten training runs were averaged to determine the influence of the random selection of annotators. For the last two experiments (3) and (4) the annotators were sorted in descending order by their label agreement compared to the majority vote measured by the F$_1$~score. The model was then trained on (3) the first n pathologists ($\hat{=}$ highest agreement) and (4) on the last n pathologists ($\hat{=}$ lowest agreement). 

By using the consensus, we only trained with mitotic figures with at least 50~\% agreement. We modified the loss computation by weighting the sample with the percentage of pathologists that agreed upon the respective sample ($>$~50-100~\%).

\subsection{Model evaluation}
The test set was generated from ten test slides and comprised all mitotic figure labels upon which the majority of all twelve pathologists (i.e. at least seven pathologists) agreed. For model performance evaluation, the F$_1$~score was computed. As previously defined by Bertram et al.~\cite{3023-06}, a mitotic figure detection was counted as true positive if the Euclidean distance of the reference and predicted bounding box centroids was at most 25 pixels ($\hat{=}$ 6.25~${\mu m}$, i.e. approximately the average cell radius of neoplastic mast cells). The detector confidence threshold was chosen based on the highest performance on the validation set.

\section{Results}
\begin{figure}[!ht]
\setcapindent{0pt}
\centering
\includegraphics[width=1\linewidth]{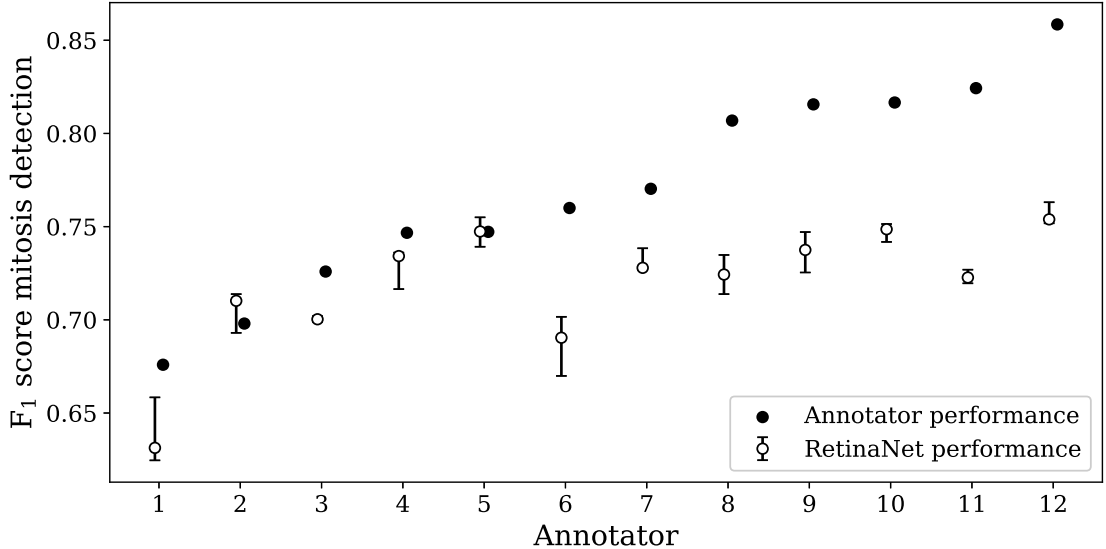}
\caption{\label{3023-fig-01}Labelling performance (F$_1$~score) of individual annotator compared to the performance of networks trained with annotations from each respective pathologist (median and range of three training runs) evaluated on the majority vote test set.}
\end{figure}

Label accuracy of the twelve pathologists compared to the majority vote was highly variable with F$_1$~scores ranging from 0.64 to 0.85 (median: 0.77) for the whole dataset and from 0.68 to 0.86 (median: 0.77) for the test cases. In Fig.~\ref{3023-fig-01} the labelling performance of each annotator is compared to the F$_1$~scores of the RetinaNet, which was trained on mitotic figure labels of the same pathologists. For the five pathologists with the lowest label agreement on the reference annotations, the algorithmic approach showed similar performance on the test dataset. Regardless of the label performance of the annotators, the algorithmic F$_1$~score was capped at around 0.74 for the network architecture used.

\begin{figure}[!ht]
\setcapindent{0pt}
\centering
\includegraphics[width=1\linewidth]{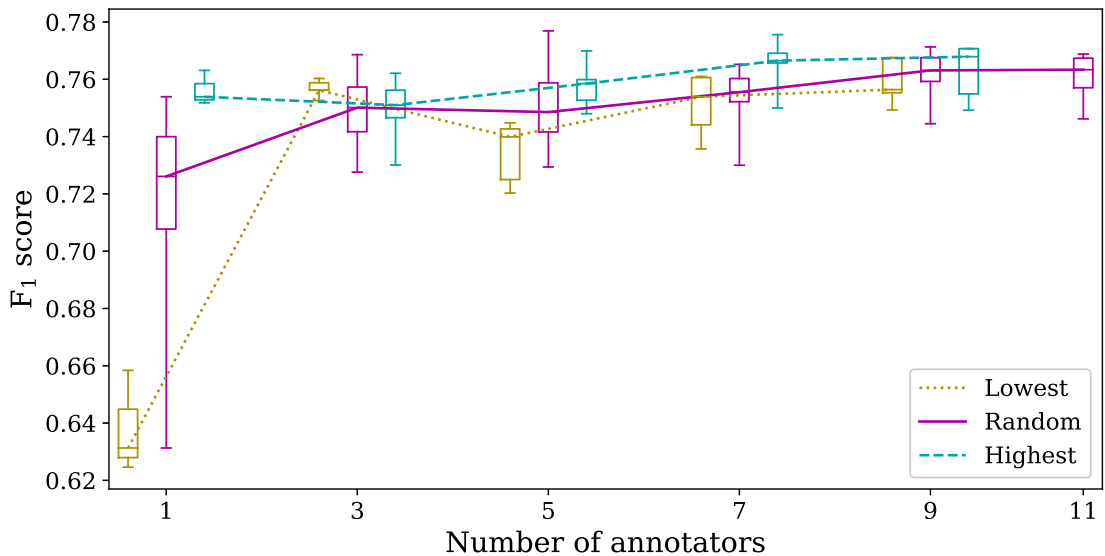}
\caption{\label{3023-fig-02}Performance (F$_1$~score) of models trained with different combinations of annotators. Whiskers represent the minimum and maximum F$_1$~score from the training runs.}
\end{figure}

Tab.~\ref{3023-tab-results} summarises the network performance when training the algorithm with databases of an increasingly larger subset of annotators. Compared to a single annotator, models trained with multi-expert labels resulted in an overall higher performance and a lower variability as measured by the interquartile range. Fig.~\ref{3023-fig-02} shows that training with databases that have the highest agreement with the majority vote yielded the best results. However, this arrangement benefited the least from higher numbers of annotators (as opposed to using the databases with the lowest agreement).

\begin{table}[!ht]
\setcapindent{0pt}
\centering
\caption{\label{3023-tab-results}Comparison of F$_1$~scores of models trained with different reference annotations based upon the majority vote from different combinations of annotators.}
\begin{tabular*}{\textwidth}{l@{\extracolsep\fill}cccccc}
\hline
Database  & Number of  & Training  & Median  & Min. & Max. & Interquartile \\ 
 & annotators & Runs &  &  &  & Range \\
\hline
Each once & 1& 18 x 3 & 0.726 & 0.625 & 0.763 &  0.038\\
\hline
Random & 3 & 10 & 0.75 & 0.728 & 0.769 & 0.016\\ 
Random & 5 & 10 & 0.749 & 0.729 & 0.777 &  0.017\\
Random & 7 & 10 & 0.756 & 0.730 & 0.765 &  0.008\\ 
Random & 9 & 10 & 0.763 & 0.745 & 0.771 &  0.008\\ 
Random & 11 & 10 & 0.763 & 0.746 & 0.769 &  0.010\\
\hline
Highest agreement & 3 & 5 & 0.751 & 0.73 & 0.762 & 0.010 \\ 
Highest agreement & 5 & 5 & 0.759 & 0.748 & 0.770 & 0.007 \\
Highest agreement & 7 & 5 & 0.767 & 0.75 & 0.776 & 0.003 \\ 
Highest agreement & 9 & 5 & 0.768 & 0.749 & 0.771 & 0.016 \\
\hline
Lowest agreement & 3 & 5 & 0.756 & 0.753 & 0.76 & 0.003 \\ 
Lowest agreement & 5 & 5 & 0.74 & 0.720 & 0.745 &  0.018 \\
Lowest agreement & 7 & 5 & 0.754 & 0.736 & 0.761 & 0.017 \\ 
Lowest agreement & 9 & 5 & 0.756 & 0.749 & 0.768 & 0.012 \\ 
\hline
\end{tabular*}
\end{table}

\section{Discussion}
The high variability between annotators in the present study is consistent with previous studies~\cite{3023-02,3023-03,3023-07} and warrants a detailed evaluation of label consistency and the impact on algorithmic performance, as was the goal of the present study. Generally, our results show that the use of a consensus of a higher number of pathologists for training the algorithm yields better and more consistent results. In particular, F$_1$~scores were noticeably variable when training the algorithm with single annotators (interquartile range: 0.038). A consensus by three pathologists was highly beneficial for more consistent training results (interquartile range: 0.016). Increasing the number of randomly selected annotators further only improved the median F$_1$~score by a small amount (+~0.013). Nevertheless, we have also shown that even training with a single annotator can result in high performance. Our results emphasise that high annotation accuracy and consensus by a small number of pathologists may result in the best trade-off between algorithmic performance and labour intensity of dataset development. Further enhancement of label consistency may be achieved with repeated screening of images or algorithmically augmented labelling~\cite{3023-06}. An interesting approach to reduce subjectivity for future mitotic figures datasets was recently introduced by Tellez et al.~\cite{3023-08}. In this study, a specific immunohistochemical marker of mitotic figures was used to derive object labels and labels were assigned to images with standard histologic stain via image registration. 

The major limitation of the conducted experiments was the pathologist-defined ground truth. Although pathologists are the current gold standard for labelling histologic images, they have high inter-rater variability which hampers not only training of data-driven algorithms (as proven in the present study) but also biases performance evaluation. The finding that algorithms trained with multi-expert databases outperformed many pathologists on the test set was attributed to the fact that algorithms yielded high sensitivity. Furthermore, the experiments in the present work were limited to a standard object detection architecture with relatively low complexity. Compensation for noisy labels and higher F$_1$~scores may be achieved with a more complex model, such as by adding a second classification stage~\cite{3023-01,3023-06}, and larger training datasets. Further improvement of data-derived algorithms may be accomplished with advanced deep learning methods that incorporate label accuracy during training.

\section*{Acknowledgement}
FW gratefully acknowledges financial support received by Merck KGaA. CAB gratefully acknowledges financial support received from the Dres. Jutta \& Georg Bruns-Stiftung f\"ur innovative Veterin\"armedizin.

\bibliographystyle{bvm}

\bibliography{3023}

\end{document}